\documentclass{article}
\usepackage[T1]{fontenc} 
\usepackage[utf8]{inputenc} 
\usepackage{ismir,amsmath,cite,url}
\usepackage{graphicx}
\usepackage{color}

\usepackage{lineno}

\title{Composer Style Classification of Piano Sheet Music Images Using Language Model Pretraining}

\twoauthors
  {TJ Tsai} {Harvey Mudd College \\ {\tt ttsai@g.hmc.edu}}
  {Kevin Ji} {Harvey Mudd College \\ {\tt kji@g.hmc.edu}}

\begin{document}

\maketitle
\begin{abstract}
This paper studies composer style classification of piano sheet music images.  Previous approaches to the composer classification task have been limited by a scarcity of data.  We address this issue in two ways: (1) we recast the problem to be based on raw sheet music images rather than a symbolic music format, and (2) we propose an approach that can be trained on unlabeled data.  Our approach first converts the sheet music image into a sequence of musical ``words" based on the bootleg feature representation, and then feeds the sequence into a text classifier.  We show that it is possible to significantly improve classifier performance by first training a language model on a set of unlabeled data, initializing the classifier with the pretrained language model weights, and then finetuning the classifier on a small amount of labeled data.  We train AWD-LSTM, GPT-2, and RoBERTa language models on all piano sheet music images in IMSLP.  We find that transformer-based architectures outperform CNN and LSTM models, and pretraining boosts classification accuracy for the GPT-2 model from 46\% to 70\% on a 9-way classification task.  The trained model can also be used as a feature extractor that projects piano sheet music into a feature space that characterizes compositional style.
\end{abstract}
\section{Introduction}
\label{sec:introduction}

We've all had the experience of hearing a piece of music that we've never heard before, but immediately recognizing the composer based on the piece's style.  This paper explores this phenomenon in the context of sheet music.  The question that we want to answer is: ``Can we predict the composer of a previously unseen page of piano sheet music based on its compositional style?"

Many previous works have studied the composer classification problem.  These works generally fall into one of two categories.  The first category of approach is to construct a set of features from the music, and then feed the features into a classifier.  Many works use manually designed features that capture musically meaningful information (e.g. \cite{kempfert2018does}\cite{sadeghian2017classification}\cite{brinkman2016musical}\cite{herremans2016composer}).  Other works feed minimally preprocessed representations of the data (e.g. 2-D piano rolls \cite{velarde2018convolution}\cite{velarde2016composer} or tensors encoding note pitch \& duration information \cite{thickstun2019midi}\cite{buzzanca2002supervised}) into a convolutional model, and allow the model to learn a useful feature representation.  The second category of approach is to train one model for each composer, and then select the model that has the highest likelihood of generating a given sequence of music.  Common approaches in this category include N-gram language models \cite{wolkowicz2013evaluation}\cite{hontanilla2013modeling}\cite{hillewaere2010string} and Markov models \cite{kaliakatsos2011weighted}\cite{pollastri2001classification}.

Our approach to the composer classification task addresses what we perceive to be the biggest common obstacle to the above approaches: lack of data.  All of the above approaches assume that the input is in the form of a symbolic music file (e.g. MIDI or **kern).  Because symbolic music formats are much less widely used than audio, video, and image formats, the amount of training data that is available is quite limited.  We address this issue of data scarcity in two ways: (1) we re-define the composer classification task to be based on sheet music images, for which there is a lot of data available online, and (2) we propose an approach that can be trained on unlabeled data.

Our work takes advantage of recent developments in transfer learning in the natural language processing (NLP) community.  Prior to 2017, transfer learning in NLP was done in a limited way.  Typically, one would use pretrained word embeddings such as word2vec \cite{mikolov2013distributed}\cite{mikolov2013efficient} or GloVe \cite{pennington2014glove} vectors as the first layer in a model.  The problem with this paradigm of transfer learning is that the entire model except the first layer needs to be trained from scratch, which requires a large amount of labeled data.  This is in contrast to the paradigm of transfer learning in computer vision, where a model is trained on the ImageNet classification task \cite{russakovsky2015imagenet}, the final layer is replaced with a different linear classifier, and the model is finetuned for a different task.  The benefit of this latter paradigm of transfer learning is that the entire model except the last layer is pretrained, so it can be finetuned with only a small amount of labeled data.  This paradigm of transfer learning has been widely used in computer vision in the last decade \cite{yosinski2014transferable} using pretrained models like VGG \cite{simonyan2014very}, ResNet \cite{he2016deep}, Densenet \cite{huang2017densely}, etc.  The switch to ImageNet-style transfer learning in the NLP community occurred in 2017, when Howard et al. \cite{howard2018universal} proposed a way to pretrain an LSTM-based language model on a large set of unlabeled data, add a classification head on top of the language model, and then finetune the classifier on a new task with a small amount of labeled data.  This was quickly followed by several other similar language model pretraining approaches that replaced the LSTM with transformer-based architectures (e.g. GPT \cite{radford2018improving}, GPT-2 \cite{radford2019language}, BERT \cite{devlin2018bert}).  These pretrained language models have provided the basis for achieving state-of-the-art results on a variety of NLP tasks, and have been extended in various ways (e.g. Transformer-XL \cite{dai2019transformer}, XLNet \cite{yang2019xlnet}).

Our approach is similarly based on language model pretraining.  We first convert each sheet music image into a sequence of words based on the bootleg score feature representation \cite{yang2019midi}.  We then feed this sequence of words into a text classifier.  We show that it is possible to significantly improve the performance of the classifier by training a language model on a large set of unlabeled data, initialize the classifier with the pretrained language model weights, and finetune the classifier on a small amount of labeled data.  In our experiments, we train language models on all piano sheet music images in the International Music Score Library Project (IMSLP)\footnote{\url{http://imslp.org/}} using the AWD-LSTM \cite{merity2017regularizing}, GPT-2 \cite{radford2019language}, and RoBERTa \cite{liu2019roberta} language model architectures.  By using pretraining, we are able to improve the accuracy of our GPT-2 model from 46\% to 70\% on a 9-way classification task.\footnote{Code can be found at \url{https://github.com/tjtsai/PianoStyleEmbedding}.}


\begin{figure}
	\centerline{\includegraphics[width=\columnwidth]{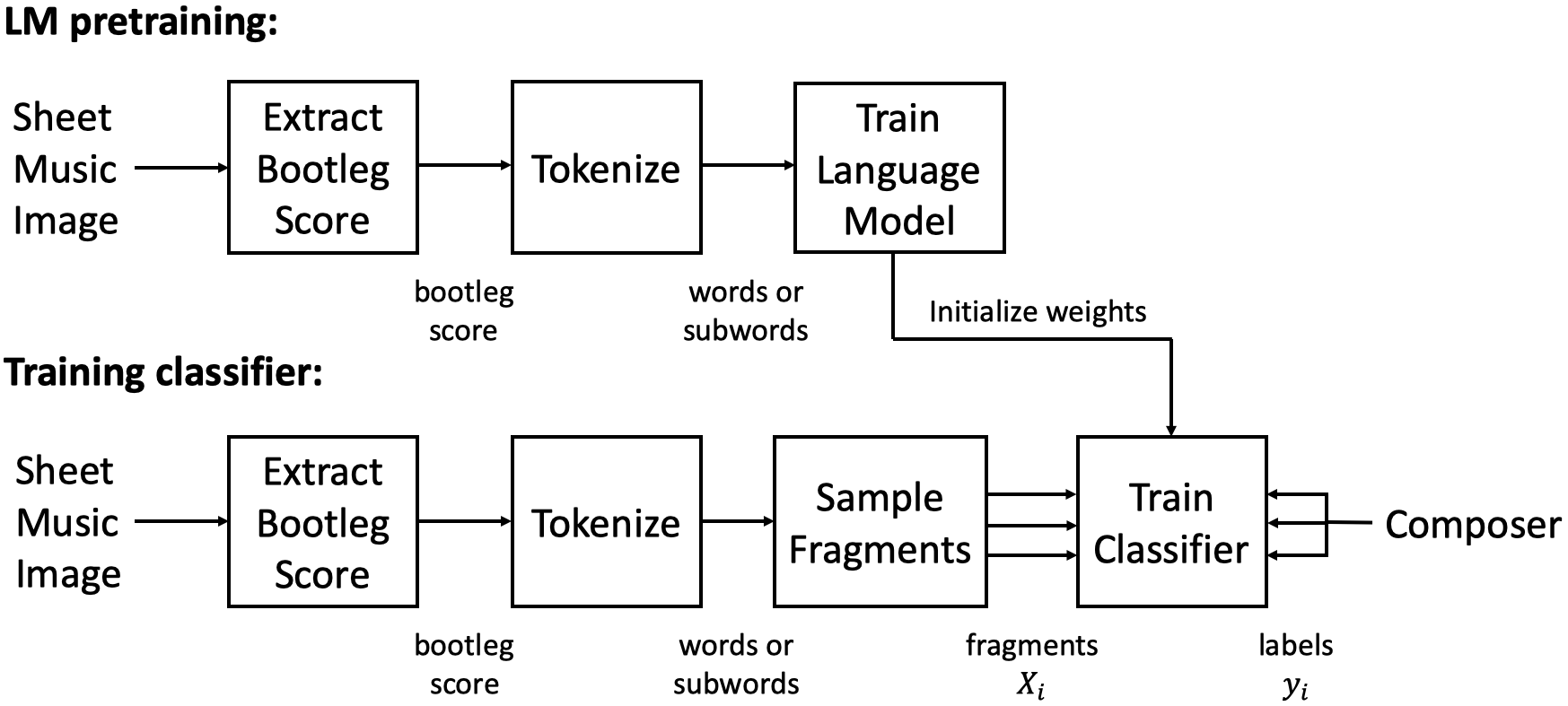}}
	\caption{Overview of proxy classifier training.  A language model is first trained on a set of unlabeled data, the classifier is initialized with the pretrained language model weights, and then the classifier is finetuned on a small set of labeled data.}
	\label{fig:trainingOverview}
\end{figure}

\section{System Description}
\label{sec:systemDescr}

We will describe our system in the next four subsections.  In the first subsection, we give a high-level overview and rationale behind our approach.  In the following three subsections, we describe the three main stages of system development: language model pretraining, classifier finetuning, and inference.

\subsection{Overview}
\label{subsec:overview}

Figure \ref{fig:trainingOverview} summarizes our training approach.  In the first stage, we convert each sheet music image into a sequence of words based on the bootleg score representation \cite{yang2019midi}, and then train a language model on these words.  Since this task does not require labels, we can train our language model on a large set of unlabeled data.  In this work, we train our language model on all piano sheet music images in the IMSLP dataset.  In the second stage, we train a classifier that predicts the composer of a short fragment of music, where the fragment is a fixed-length sequence of symbolic words.  We do this by adding one or more dense layers on top of the language model, initializing the weights of the classifier with the language model weights, and then finetuning the model on a set of labeled data.  In the third stage, we use the classifier to predict the composer of an unseen scanned page of piano sheet music.  We do this by converting the sheet music image to a sequence of symbolic words, and then either (a) applying the classifier to a single variable length input sequence, or (b) averaging the predictions of fixed-length crops sampled from the input sequence.  We will describe each of these three stages in more detail in the following three subsections.

\begin{figure}
	\centerline{\includegraphics[width=\columnwidth]{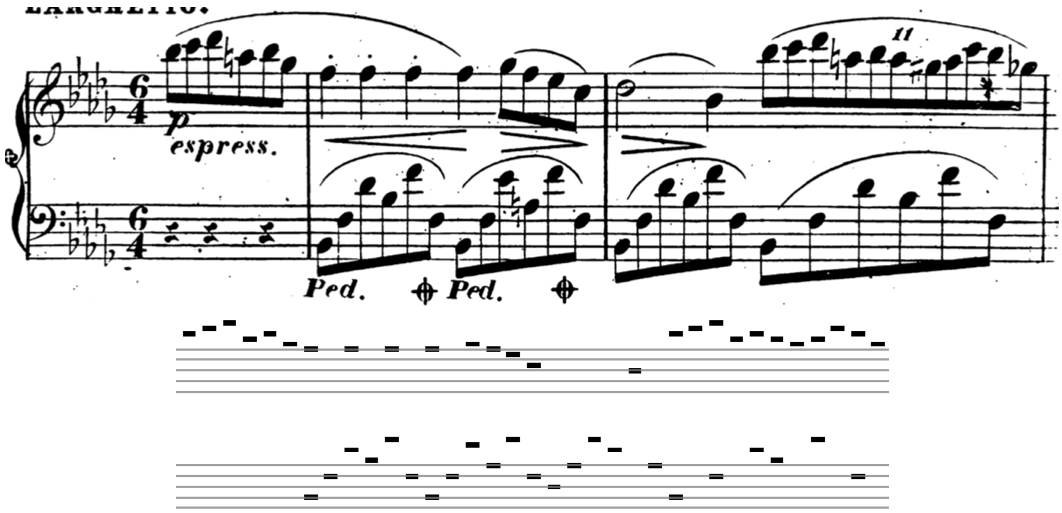}}
	\caption{A short section of sheet music and its corresponding bootleg score.  Staff lines in the bootleg score are shown for reference, but are not present in the actual feature representation.}
	\label{fig:exampleBootlegScore}
\end{figure}

The guiding principle behind our approach is to maximize the amount of data.  This impacts our approach in three significant ways.  First, it informs our choice of data \textit{format}.  Rather than using symbolic scores (as in previous approaches), we instead choose to use raw sheet music images.  While this arguably makes the task much more challenging, it has the benefit of having much more data available online.  Second, we choose an approach that can utilize unlabeled data.  Whereas labeled data is usually expensive to annotate and limited in quantity, unlabeled data is often extremely cheap and available in abundance.  By adopting an approach that can use unlabeled data, we can drastically increase the amount of data available to train our models.  Third, we use data augmentation to make the most of the limited quantity of labeled data that we do have.  Rather than fixating on the page classification task, we instead define a proxy task where the goal is to predict the composer given a fixed-length sequence of symbolic words.  By defining the proxy task in this way, we can aggressively subsample fragments from the labeled data, resulting in a much larger number of unique training data points than there are actual pages of sheet music.  Once the proxy task classifier has been trained, we can apply it to the full page classification task in a straightforward manner.

\subsection{Language Model Pretraining}
\label{subsec:pretraining}

The language model pretraining consists of three steps, as shown in the upper half of Figure \ref{fig:trainingOverview}.  These three steps will be described in the next three paragraphs.

\begin{figure}
	\centerline{\includegraphics[width=\columnwidth]{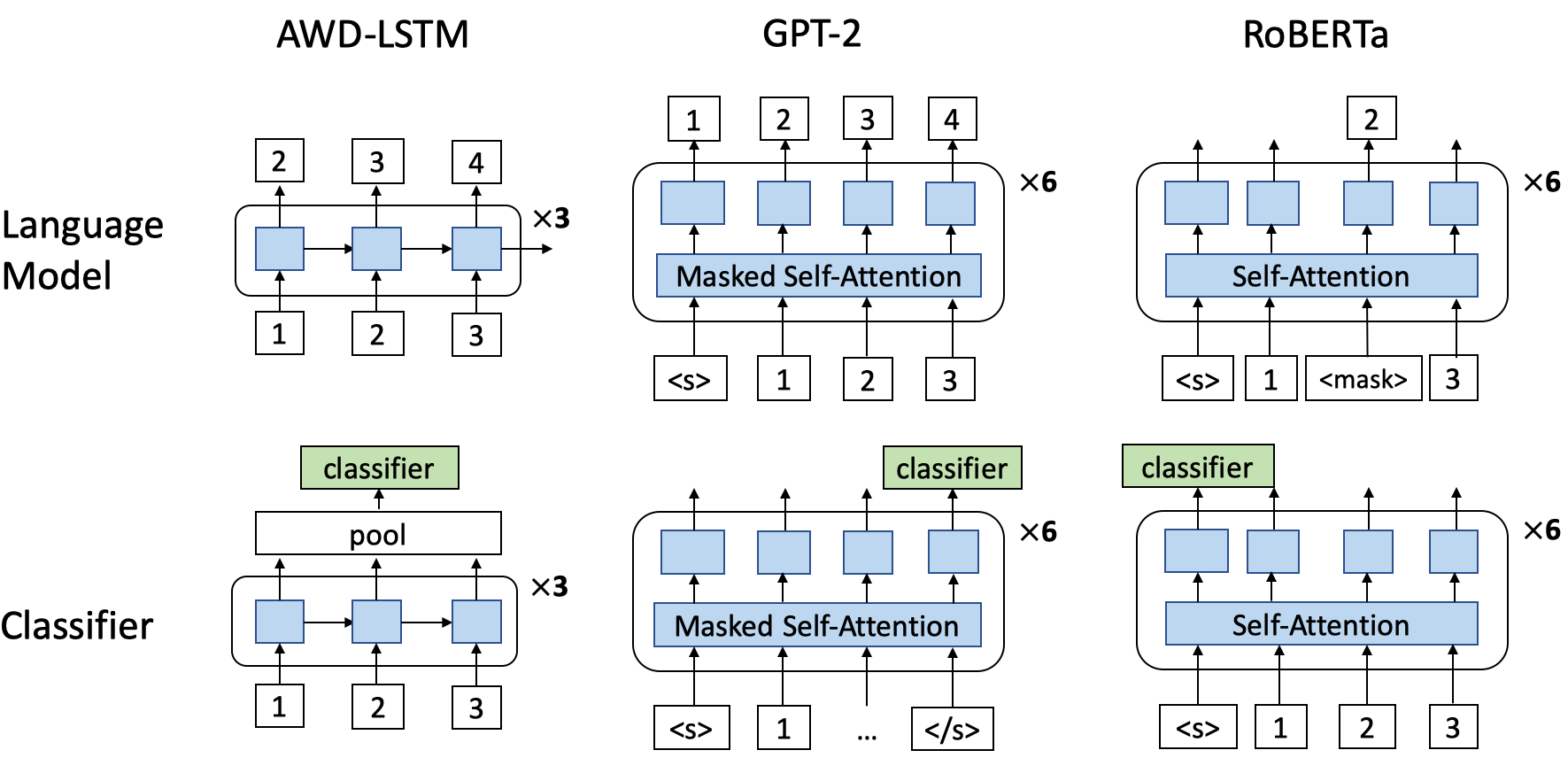}}
	\caption{Overview of AWD-LSTM, GPT-2, and RoBERTa language models (top) and classifiers (bottom).  Boxes in blue are trained during the language modeling phase and used to initialize the classifier.}
	\label{fig:modelOverview}
\end{figure}

The first step is to convert the sheet music image into a bootleg score.  The bootleg score is a low-dimensional feature representation of piano sheet music that encodes the position of filled noteheads relative to the staff lines \cite{yang2019midi}.  Figure \ref{fig:exampleBootlegScore} shows an example of a section of sheet music and its corresponding bootleg score representation.  The bootleg score itself is a $62 \times N$ binary matrix, where $62$ indicates the total number of possible staff line positions in both the left and right hands, and where $N$ indicates the total number of estimated simultaneous note onset events.  Note that the representation discards a significant amount of information: it does not encode note duration, key signature, time signature, measure boundaries, accidentals, clef changes, or octave markings, and it simply ignores non-filled noteheads (e.g. half or whole notes).  Nonetheless, it has been shown to be effective in aligning sheet music and MIDI \cite{yang2019midi}, and we hypothesize that it may also be useful in characterizing piano style.  The main benefit of using the bootleg score representation over a full optical music recognition (OMR) pipeline is processing time: computing a bootleg score only takes about $1$ second per page using a CPU, which makes it suitable for computing features on the entire IMSLP dataset.\footnote{In contrast, the best performing music object detectors take 40-80 seconds to process each page at inference time using a GPU \cite{pacha2018baseline}.}  We use the code from \cite{yang2019midi} as a fixed feature extractor to compute the bootleg scores.

The second step is to tokenize the bootleg score into a sequence of word or subword units.  We do this differently for different language models.  For word-based language models (e.g. AWD-LSTM \cite{merity2017regularizing}), we consider each bootleg score column as a single word consisting of a 62-character string of 0s and 1s.  We limit the vocabulary to the $30,000$ most frequent words, and map infrequent words to a special unknown word token <unk>.  For subword-based language models (e.g. GPT-2 \cite{radford2019language}, RoBERTa \cite{liu2019roberta}), we use a byte pair encoding (BPE) algorithm \cite{gage1994new} to learn a vocabulary of subword units in an unsupervised manner.  The BPE algorithm starts with an initial set of subword units (e.g. the set of unique characters \cite{sennrich2015neural} or the $2^8=256$ unique byte values that comprise unicode characters \cite{gillick2016multilingual}), and it iteratively merges the most frequently occurring pair of adjacent subword units until a desired vocabulary size has been reached.  We experimented with both character-level and byte-level encoding schemes (i.e. representing each word as a string of 62 characters vs. a sequence of 8 bytes), and we found that the byte-level encoding scheme performs much better.  We only report results with the byte-level BPE tokenizer.  For both subword-based language models explored in this work, we use the same shared BPE tokenizer with a vocabulary size of $30,000$ (which is the vocabularly size used in the RoBERTa model).  At the end of the second step, we have represented the sheet music image as a sequence of words or subword units.

The third step is to train a language model on a set of unlabeled data.  In this work, we explore three different language models, which are representative of state-of-the-art models in the last 3-4 years.  The top half of Figure \ref{fig:modelOverview} shows a high-level overview of these three language models.  The first model is AWD-LSTM \cite{merity2017regularizing}.  This is a 3-layer LSTM architecture that makes heavy use of regularization techniques throughout the model, including four different types of dropout.  The output of the final LSTM layer is fed to a linear decoder whose weights are tied to the input embedding matrix.  This produces an output distribution across the tokens in the vocabulary.  The model is then trained to predict the next token at each time step.  We use the fastai implementation of the AWD-LSTM model with default parameters.  The second model is openAI's GPT-2 \cite{radford2019language}.  This architecture consists of multiple transformer decoder layers \cite{vaswani2017attention}.  Each transformer decoder layer consists of a masked self-attention, along with feedforwards layers, layer normalizations, and residual connections.  While transformer encoder layers allow each token to attend to all other tokens in the input, the transformer decoder layers only allow a token to attend to previous tokens.\footnote{This is because, in the original machine translation task \cite{vaswani2017attention}, the decoder generates the output sentence autoregressively.}  Similar to the AWD-LSTM model, the outputs of the last transformer layer are fed to a linear decoder whose weights are tied to the input embeddings, and the model is trained to predict the next token at each time step.  We use the huggingface implementation of the GPT-2 model with default parameters, except that we reduce the vocabulary size from $50,000$ to $30,000$ (to use the same tokenizer as the RoBERTa model), the amount of context from 1024 to 512, and the number of layers from 12 to 6.  The third model is RoBERTa \cite{liu2019roberta}, which is based on Google's BERT language model \cite{devlin2018bert}.  This architecture consists of multiple transformer encoder layers.  Unlike GPT-2, each token can attend to all other tokens in the input and the goal is not to predict the next token.  Instead, a certain fraction of the input tokens are randomly converted to a special <mask> token, and the model is trained to predict the masked tokens.  We use the huggingface implementation of RoBERTa with default parameter settings, except that we reduce the number of layers from 12 to 6.

\subsection{Classifier Finetuning}
\label{subsec:finetuning}

In the second main stage, we finetune a classifier based on a set of labeled data.  The labeled data consists of a set of sheet music images along with their corresponding composer labels.  The process of training the classifier is comprised of four steps (lower half of Figure \ref{fig:trainingOverview}).

The first two steps are to compute and tokenize a bootleg score into a sequence of symbolic words.  We use the same fixed feature extractor and the same tokenizer that were used in the language model pretraining stage.

The third step is to sample short, fixed-length fragments of words from the labeled data.  As mentioned in Section \ref{subsec:overview}, we define a proxy task where the goal is to predict the composer given a short, fixed-length fragment of words.  Defining the proxy task in this way has three significant benefits: (1) we can use sampling to generate many more unique training data points than there are actual pages of sheet music in our dataset, (2) we can sample the data in such a way that the classes are balanced, which avoids problems during training, and (3) using fixed-length inputs allows us to train more efficiently in batches.  Our approach follows the general recommendations of a recent study on best practices for training a classifier with imbalanced data \cite{buda2018systematic}.  Each sampled fragment and its corresponding composer label constitute a single $(X_i, y_i)$ training pair for the proxy task.

The fourth step is to train the classifier model.  The bottom half of Figure \ref{fig:modelOverview} shows how this is done with our three models.  Our general approach is to add a classifier head on top of the language model, initialize the weights of the classifier with the pretrained language model weights, and then finetune the classifier on the proxy task data.  For the AWD-LSTM, we take the outputs from the last LSTM layer and construct a fixed-size representation by concatenating three things: (a) the output at the last time step, (b) the result of max pooling the outputs across the sequence dimension, and (c) the result of average pooling the outputs across the sequence dimension.  This fixed-size representation (which is three times the hidden dimension size) is then fed into the classifier head, which consists of two dense layers with batch normalization and dropout.  For the GPT-2 model, we take the output from the last transformer layer at the last time step, and then feed it into a single dense (classification) layer.  Because the GPT-2 and RoBERTa models require special tokens during training, we insert special symbols <s> and </s> at the beginning and end of every training input, respectively.  Because of the masked self-attention, we must use the output of the last token in order to access all of the information in the input sequence.  For the RoBERTa model, we take the output from the last transformer layer corresponding to the <s> token, and feed it into a single dense (classification) layer.  The <s> takes the place of the special [CLS] token described in the original paper.

We integrated all models into the fastai framework and finetuned the classifier in the following manner.  We first select an appropriate learning rate using a range test, in which we sweep the learning rate across a wide range of values and observe the impact on training loss.  We initially freeze all parameters in the model except for the untrained classification head, and we gradually unfreeze more and more layers in the model as the training converges.  To avoid overly aggressive changes to the pretrained language model weights, we use discriminative finetuning, in which earlier layers of the model use exponentially smaller learning rates compared to later layers in the model.  All training is done with (multiple cycles of) the one cycle training policy \cite{smith2018disciplined}, in which learning rate and momentum are varied cyclically over each cycle.  The above practices were proposed in \cite{howard2018universal} and found to be effective in finetuning language models for text classification.

\subsection{Inference}
\label{subsec:inference}

The third main stage is to apply the proxy classifier to the original full page classification task.  We explore two different ways to do this.  The first method is to convert the sheet music image into a bootleg score, tokenize the bootleg score into a sequence of word or subword units, and then apply the proxy classifier to a single variable-length input.  Note that all of the models can handle variable-length inputs up to a maximum context length.  The second method is identical to the first, except that it averages the predictions from multiple fixed-length crops taken from the input sequence.  The fixed-length crops are the same size as is used during classifier training, and the crops are sampled uniformly with $50\%$ overlap.\footnote{We also experimented with applying a Bayesian prior to the classifier softmax outputs, as recommended in \cite{buda2018systematic}, but found that the results were not consistently better.}

\section{Experimental Setup}
\label{sec:setup}

In this section we describe the data collection process and the metrics used to evaluate our approach.

\begin{figure}
	\centerline{\includegraphics[width=\columnwidth]{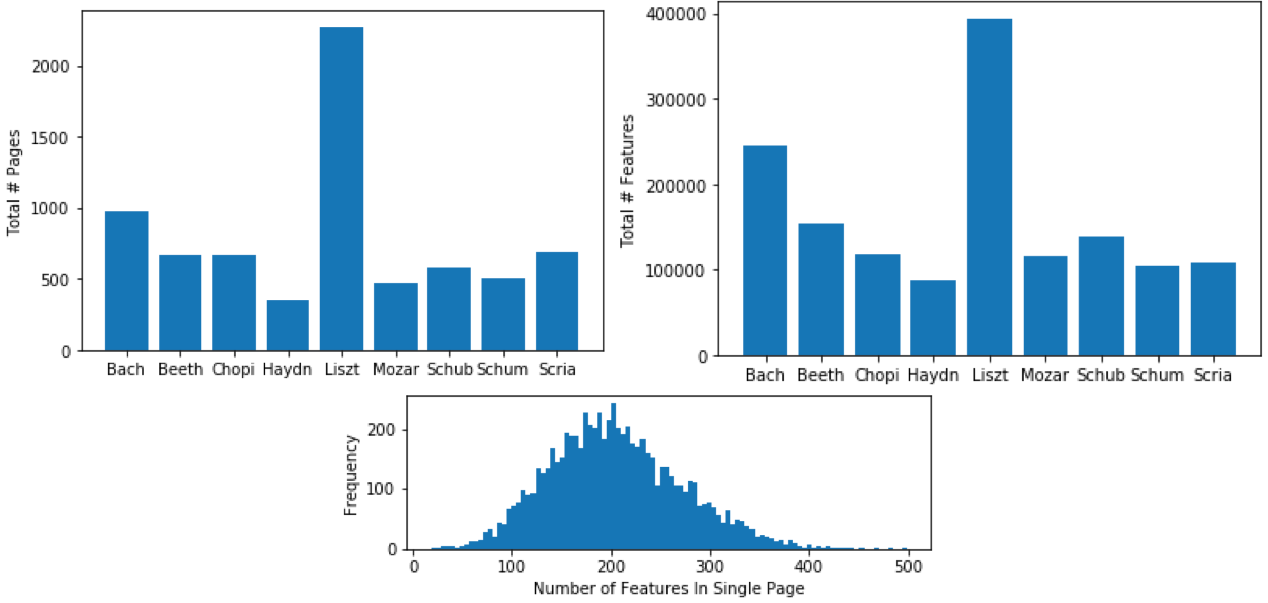}}
	\caption{Statistics on the target dataset.  The top two histograms show the distribution of the number of pages (top left) and number of bootleg score features (top right) per composer.  The bottom figure shows the distribution of the number of bootleg score features per page.}
	\label{fig:dataStats}
\end{figure}

The data comes from IMSLP.  We first scraped the website and downloaded all PDF scores and accompanying metadata.\footnote{We downloaded the data over a span of several weeks in May of 2018.}  We filtered the data based on its instrumentation in order to identify a list of solo piano scores.  We then computed bootleg score features for all of the piano sheet music images using the XSEDE supercomputing infrastructure \cite{xsede}, and discarded any pages that had less than a minimum threshold of features.  This latter step is designed to remove non-music pages such as the title page, foreword, or table of contents.  The resulting set of data contained $29,310$ PDFs,\footnote{Note that a PDF may contain multiple pieces (e.g. the complete set of Chopin etudes).} $255,539$ pages and a total of $48.5$ million bootleg score features.  This set of data is what we refer to as the IMSLP dataset in this work (e.g. the IMSLP pretrained language model).  For language model training, we split the IMSLP data by piece, using $90\%$ for training and $10\%$ for validation.

The classification task uses a subset of the IMSLP data.  We first identified a list of composers with a significant amount of data (composers shown in Figure \ref{fig:dataStats}).  We limited the list to nine composers in order to avoid extreme class imbalance.  Because popular pieces tend to have many sheet music versions in the  dataset, we select one version per piece in order to avoid over-representation of a small subset of pieces.  Next, we manually labeled and discarded all filler pages, and then computed bootleg score features on the remaining sheet music images.  This cleaned dataset is what we refer to as the target data in this work (e.g. the target pretrained language model).  Figure \ref{fig:dataStats} shows the total number of pages and bootleg score features per composer for the target dataset, along with the distribution of the number of bootleg score features per page.  For training and testing, we split the data by piece, using $60\%$ of the pieces for training ($4347$ pages), $20\%$ for validation ($1500$ pages), and $20\%$ for testing ($1304$ pages).  To generate data for the proxy task, we randomly sampled fixed-length fragments from the target data.  We sample the same number of fragments for each composer to ensure class balance.  We experimented with fragment sizes of 64/128/256 and sampled 32400/16200/8100 fragments for training and 10800/5400/2700 fragments for validation/test, respectively.  This sampling scheme ensures the same data coverage regardless of fragment length.  Note that the classification data is carefully curated, while the IMSLP data requires minimal processing.

We use two different metrics to evaluate our systems.  For the proxy task, accuracy is an appropriate metric since the data is balanced.  For the full page classification task -- which has imbalanced data -- we report results in macro F1 score.  Macro F1 is a generalization of F1 score to a multi-class setting, in which each class is treated as a one-versus-all binary classification task and the F1 scores from all classes are averaged.

\section{Results \& Analysis}
\label{sec:results}

In this section we present our experimental results and conduct various analyses to answer key questions of interest.  While the proxy task is an artificially created task, it provides a more reliable indicator of classifier performance than the full page classification.  This is because the test set of the full page classification task is both imbalanced and very small ($1304$ data points).  Accordingly, we will report results on both the proxy task and full page classification task.

\subsection{Proxy Task}

We first consider the performance of our models on the proxy classification task.  We would like to understand the effect of (a) model architecture, (b) pretraining condition, and (c) fragment size.

\begin{figure}
	\centerline{\includegraphics[width=\columnwidth]{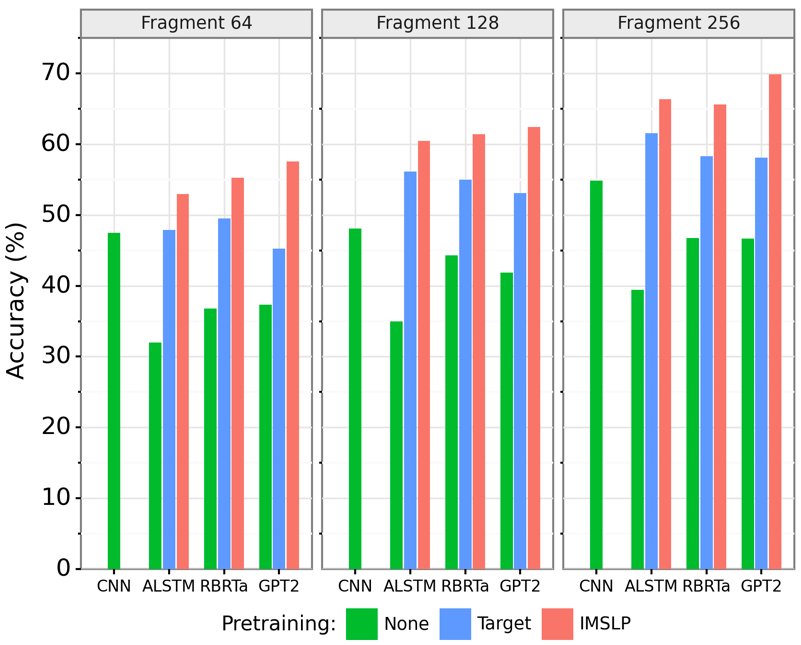}}
	\caption{Model performance on the proxy classification task.  This comparison shows the effect of different pretraining conditions and fragment sizes.}
	\label{fig:proxyResults}
\end{figure}

We evaluate four different model architectures.  In addition to the AWD-LSTM, GPT-2, and RoBERTa models previously described, we also measure the performance of a CNN-based approach recently proposed in \cite{thickstun2019midi}.  Note that we cannot use the exact same model in \cite{thickstun2019midi} since we do not have symbolic score information.  Nonetheless, we can use the same general approach of computing local features, aggregating feature statistics across time, and applying a linear classifier.  The design of our 2-layer CNN model roughly matches the architecture proposed in \cite{thickstun2019midi}.

We consider three different language model pretraining conditions.  The first condition is with no pretraining, where we train the classifier from scratch only on the proxy task.  The second condition is with target language model pretraining, where we first train a language model on the target data, and then finetune the classifier on the proxy task.  The third condition is with IMSLP language model pretraining.  Here, we train a language model on the full IMSLP dataset, finetune the language model on the target data, and then finetune the classifier on the proxy task.

Figure \ref{fig:proxyResults} shows the performance of all models on the proxy task.  There are three things to notice.  First, regarding (a), the transformer-based models generally outperform the LSTM and CNN models.   Second, regarding (b), language model pretraining improves performance significantly across the board.  Regardless of architecture, we see a large improvement going from no pretraining (condition 1) to target pretraining (condition 2), and another large improvement going from target pretraining (condition 2) to IMSLP pretraining (condition 3).  For example, the performance of the GPT-2 model increases from $37.3\%$ to $45.2\%$ to $57.5\%$ across the three pretraining conditions.  Because the data in conditions 1 \& 2 is exactly the same, the improvement in performance must be coming from more effective use of the data.  We can interpret this from an information theory perspective by noting that the classification task provides the model $log_2 9 = 3.17$ bits of information per fragment, whereas the language modeling task provides $log_2 V$ bits of information \textit{per bootleg score feature} where $V$ is the vocabulary size.  The performance gap between condition 2 and condition 3 can also be interpreted as the result of providing more information to the model, but here the information is coming from having additional data.  Third, regarding (c), larger fragments result in better performance, as we might expect.

\begin{figure}
	\centerline{\includegraphics[width=\columnwidth]{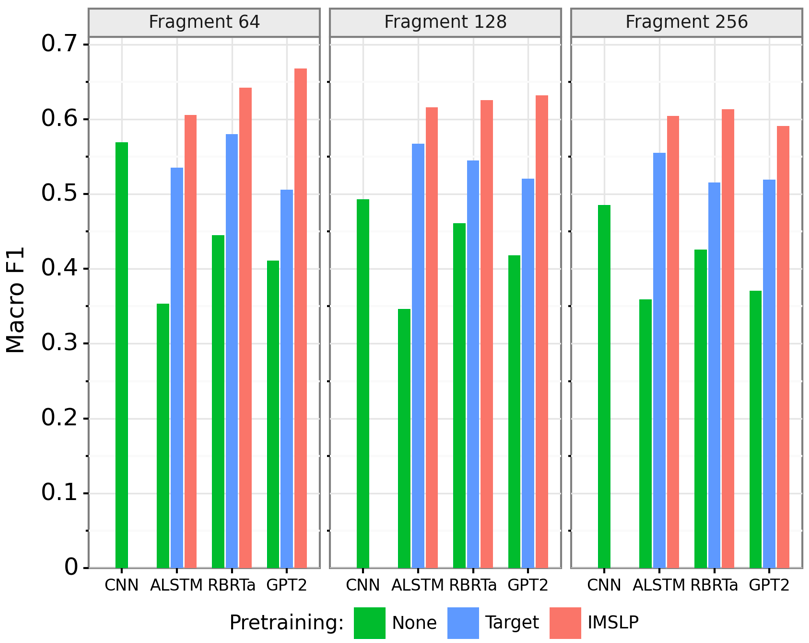}}
	\caption{Results on the full page classification task.}
	\label{fig:fullPageResults}
\end{figure}

\subsection{Full Page Classification}

Next, we consider performance of our models on the full page classification task.  We would like to understand the effect of (a) model architecture, (b) pretraining condition, (c) fragment size, and (d) inference type (single vs. multi-crop).  Regarding (d), we found that taking multiple crops improved results with all models except the CNN.  This suggests that this type of test time augmentation does not benefit approaches that simply average feature statistics over time.  In the results presented below, we only show the optimal inference type for each model architecture (i.e. CNN with single crop, all others with multi-crop).

Figure \ref{fig:fullPageResults} shows model performance on the full page classification task.  There are two things to notice.  First, we see the same general trends as in Figure \ref{fig:proxyResults} for model architecture and pretraining condition: the transformer-based models generally outperform the CNN and LSTM models, and pretraining helps substantially in every case.  The macro F1 score of our best model (GPT-2 with fragment size 64) increases from $0.41$ to $0.51$ to $0.67$ across the three pretraining conditions.  Second, we see the \textit{opposite} trend as the proxy task for fragment size: smaller fragments have \textit{better} page classification performance.  This strongly indicates a data distribution mismatch.  Indeed, when we look at the distribution of the number of bootleg score features in a single page (Figure \ref{fig:dataStats}), we see that a significant fraction of pages have less than 256 features.  Because we only sample fragments that contain a complete set of 256 words, our proxy task data is biased towards longer inputs.  This leads to poor performance when the classifier is faced with short inputs, which are never seen in training.  Using a fragment size of 64 minimizes this bias.

\begin{figure}
	\centerline{\includegraphics[width=\columnwidth]{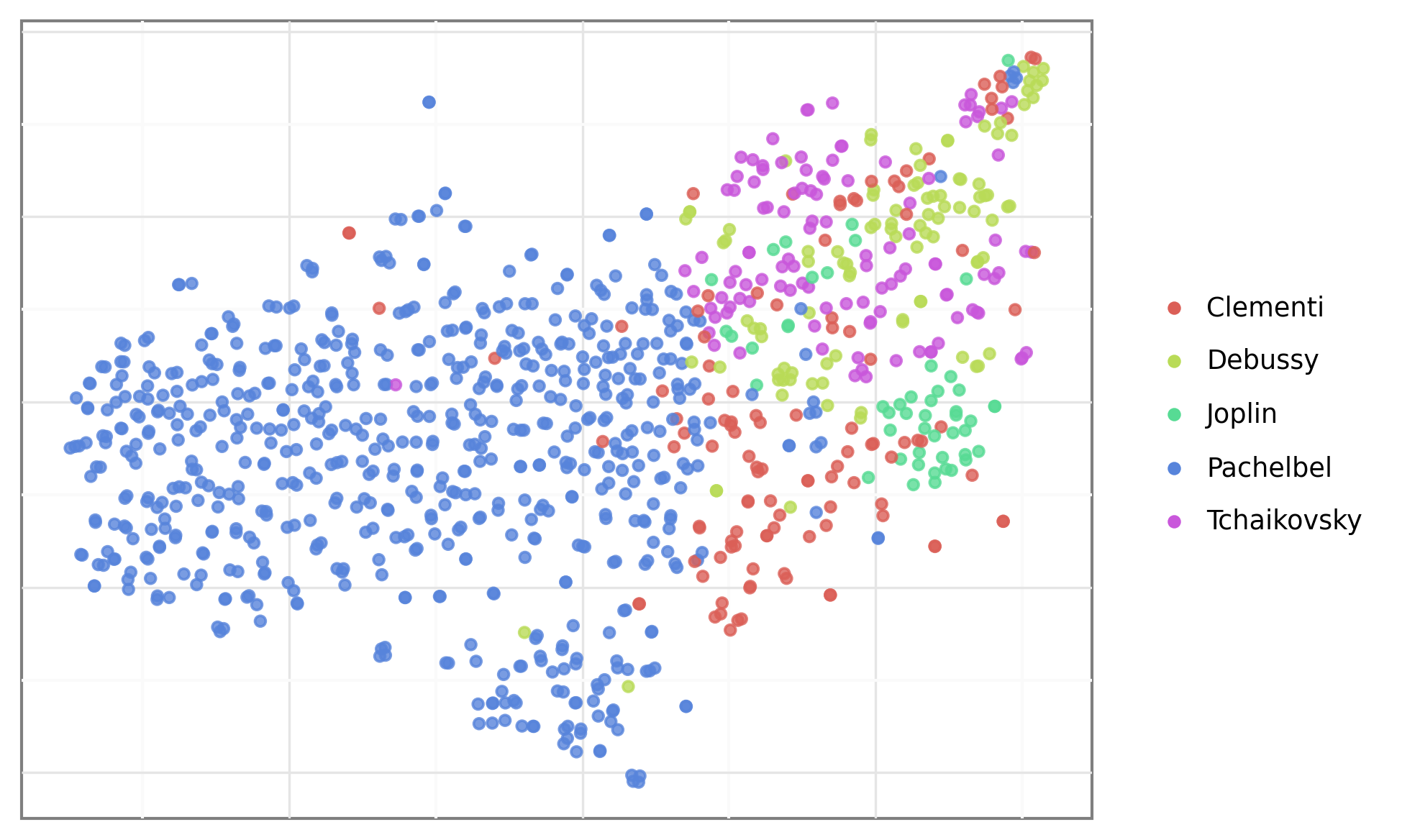}}
	\caption{t-SNE plot of the RoBERTa model activations for five novel composers.  Each data point corresponds to a single page of sheet music for a composer that was \textit{not} considered in the classification task.}
	\label{fig:tsneNovel}
\end{figure}

\subsection{t-SNE Plots}

Another key question of interest is, ``Can we use our model to characterize the style of any page of piano sheet music?"  The classification task forces the model to project the sheet music into a feature space where the compositional style of the nine composers can be differentiated.  We hypothesize that this feature space might be useful in characterizing the style of \textit{any} page of piano sheet music, even from composers not in the classification task.

To test this hypothesis, we fed data from 5 novel composers into our models and constructed t-SNE plots\cite{maaten2008visualizing} of the activations at the second-to-last layer.  Figure \ref{fig:tsneNovel} shows such a plot for the RoBERTa model.  Each data point corresponds to a single page of sheet music from a novel composer.  Even though we have not trained the classifier to distinguish between these five composers, we can see that the data points are still clustered, suggesting that the feature space can describe the style of new composers in a useful manner.

\section{Conclusion}
\label{sec:conclusion}

We propose a method for predicting the composer of a single page of piano sheet music.  Our method first converts the raw sheet music image into a bootleg score, tokenizes the bootleg score into a sequence of musical words, and then feeds the sequence into a text classifier.  We show that by pretraining a language model on a large set of unlabeled data, it is possible to significantly improve the performance of the classifier.  We also show that our trained model can be used as a feature extractor to characterize the style of any page of piano sheet music.  For future work, we would like to explore other forms of data augmentation and other model architectures that explicitly encode musical knowledge.

\section{Acknowledgments}
This work used the Extreme Science and Engineering Discovery Environment (XSEDE), which is supported by National Science Foundation grant number ACI-1548562.  Large-scale computations on IMSLP data were performed with XSEDE Bridges at the Pittsburgh Supercomputing Center through allocation TG-IRI190019.  We also gratefully acknowledge the support of NVIDIA Corporation with the donation of the GPU used for training the models.

\bibliography{PianoStyleClassification}

%
%
%
%

\end{document}